\documentclass{article}
\usepackage{spconf,amsmath,epsfig}
\usepackage{enumerate}
\usepackage{multirow}
\usepackage{subcaption}
\usepackage{color,xcolor}

\title{a Training-free, One-shot detection framework for geospatial objects in Remote Sensing Images}

\name{Tengfei Zhang$^{1,2,3}$, Yue Zhang$^{*,1,2}$\thanks{* Corresponding author.}, Xian Sun$^{1,2}$, Menglong Yan$^{1,2}$, Yaoling Wang$^{1,2,3}$, Kun Fu$^{1,2}$\thanks{ This work is supported by National Natural Science Foundation of China under Grants 41801349.}}
\address{
	1 Institute of Electronic, Chinese Academy of Sciences, Beijing, China\\
	2 Key Laboratory of Network Information System Technology (NIST), Institute of Electronics,\\ Chinese Academy of Sciences, Beijing, China\\
	3 University of Chinese Academy of Sciences, Beijing, China\\
	\tt\small zhangtengfei16@mails.ucas.ac.cn
}

\begin{document}

\maketitle

\begin{abstract}
	Deep learning based object detection has achieved great success. However, these supervised learning methods are data-hungry and time-consuming. This restriction makes them unsuitable for limited data and urgent tasks, especially in the applications of remote sensing. Inspired by the ability of humans to quickly learn new visual concepts from very few examples, we propose a training-free, one-shot geospatial object detection framework for remote sensing images. It consists of (1) a feature extractor with remote sensing domain knowledge, (2) a multi-level feature fusion method, (3) a novel similarity metric method, and (4) a 2-stage object detection pipeline. Experiments on sewage treatment plant and airport detections show that proposed method has achieved a certain effect. Our method can serve as a baseline for training-free, one-shot geospatial object detection.
\end{abstract}

\begin{figure*}[t]
	\begin{center}
		\includegraphics[width=0.9\linewidth,height=180pt]{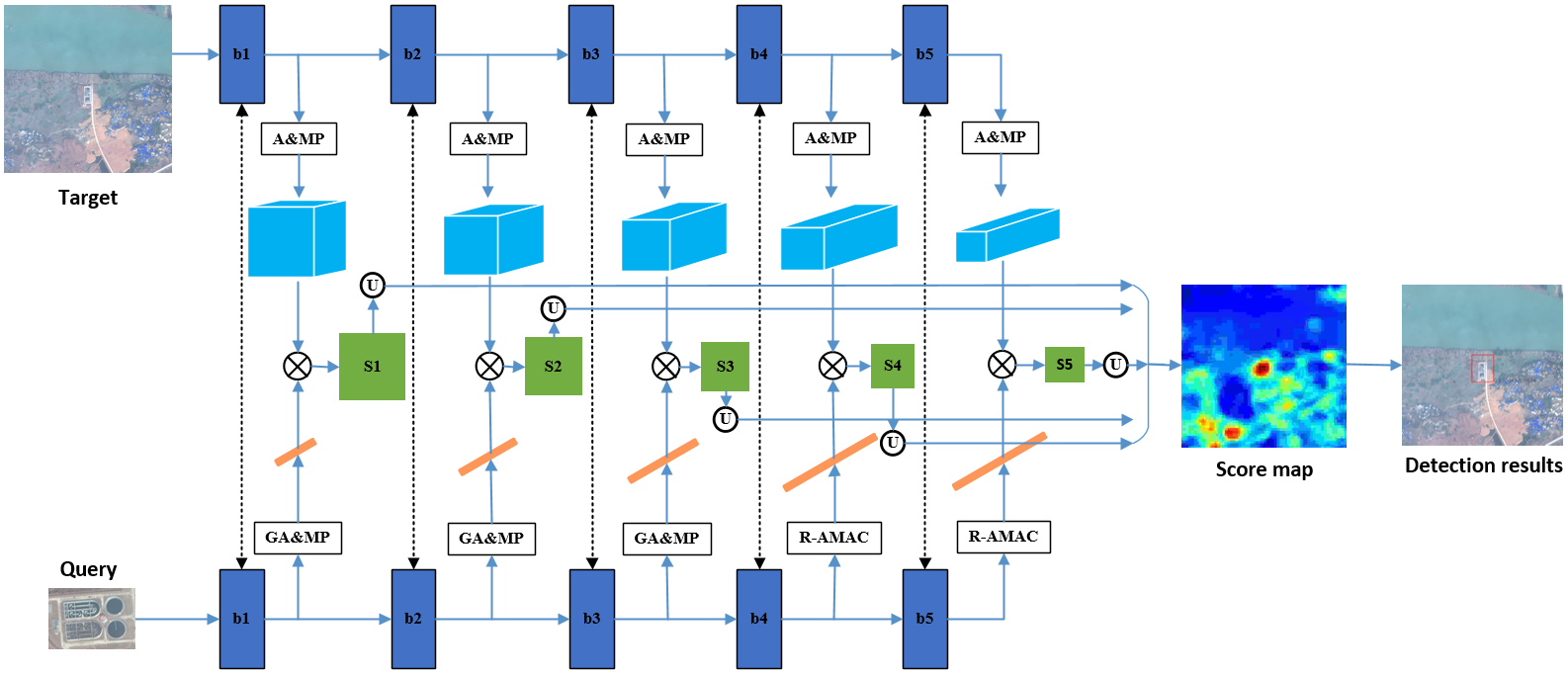}
	\end{center}
	\caption{The framework of the proposed method. The VGG16 network extracts feature maps of query and target images at each block (b1-b5). We can get 5 score maps (S1-S5) by the convolution between the query feature vectors and the target features. These score maps are up sampled to the same size as the input target image and averaged to get the final score map. Detection results can be obtained by extracting some regions of the final score map with a threshold. Here, A\&MP denotes the concatenation of the average- and max- pooling features. GA\&MP denotes the concatenation of the global average- and max- pooling features. R-AMAC (Regional Average \& Maximum Activation of Convolutions) is our proposed feature aggregation method. The $\otimes$ denotes convolution operation. The U symbol denotes upsample operation.}
	\label{fig1}
\end{figure*}\label{figure1}

\section{Introduction}
\label{sec:intro}
	Deep learning based detection methods have achieved great advances, such as Fast R-CNN \cite{Girshick2015Fast} and Faster R-CNN \cite{Ren2015Faster}. These supervised learning methods need to train models on large-scale labelled data and often take a lot of time. These inherent characteristics severely limit their performance in some cases with limited data, urgent task or dynamic environment. In contrast, humans can learn new knowledge quickly with few samples. For example, one can recognize a balance car by looking at it only once. Inspired by this ability, there are many one or few-shot learning researches on image classification, regression and reinforcement learning \cite{model_agnostic,Optimizationfewshot}, which train models from many related tasks to learn novel knowledge accurately from few labelled samples.
	
	The research on one or few-shot object detection in remote sensing is also in urgent demand. The large field and clutter background of remote sensing images make data annotation for object detection harder and more time-consuming. Due to the small sample size for geospatial objects, the performances of supervised learning methods degrade. Besides, the supervised learning systems are inflexible when faced with new object categories, because the models need to be retrained on new data.
	
	To address this issue, we propose a  training-free, one-shot geospatial object detection framework in this paper. We train the model on image classification task and test it on object detection task. As a result, it does not require any labelled data for the object detection task. More concretely, we train a feature extractor on a remote sensing image classification dataset, such that the feature extractor can learn some domain knowledge of remote sensing. Once the feature extractor has been trained, the proposed method does not need any training. Instead of training the model on the object detection task, we give it the capacity of detecting the objects of novel classes by a well-designed process. Given a small query image of a geospatial object and a large target image, the feature extractor can extract query and target features. Then we compute the similarity between query and target features to find similar regions in the target image. 

	Obviously, our method is training-free for the object detection task. Besides, this method is category-agnostic, in the sense that it can detect the objects of any categories theoretically. Our framework can get rid of the dependence on labelling object detection samples which is quite time-consuming and laborious. This can increase the flexibility and further expand the scope of application for few-shot learning. Compared with conventional supervised object detction works, settings for one-shot object detction are
	more difficult. Therefore, we implement it in a relatively simple scenario.
	Our contributions are as follows:
	\begin{enumerate}[(1)]
	\item We propose a training-free, one-shot object detection framework, which consists of a feature extractor, a multi-level feature fusion method and a novel similarity metric method.
	\item We design a 2-stage detection pipeline to improve the detection performance, and proves its effectiveness by the experiment.
	\item The proposed method has achieved a promising result for the sewage treatment plant and airport detections in remote sensing images.
	\end{enumerate}

\begin{figure}[t]
	\begin{center}
		\includegraphics[width=0.76\linewidth,height=60pt]{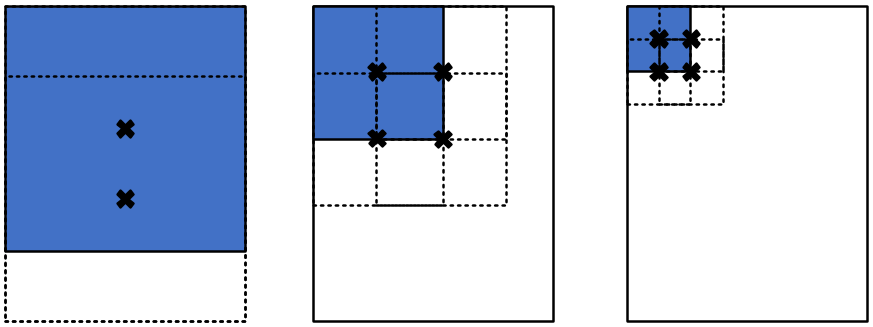}
	\end{center}
	\caption{The process of R-MAC. There are 3 different scales ($l$ = 1, 2, 3). We show the top-left region of each scale (shown in blue) and its neighboring regions with dashed borders. The cross denotes the center of a region.}
	\label{fig2}
\end{figure}\label{figure2}	

\begin{figure}[t]
	\begin{center}
		\includegraphics[width=1\linewidth,height=60pt]{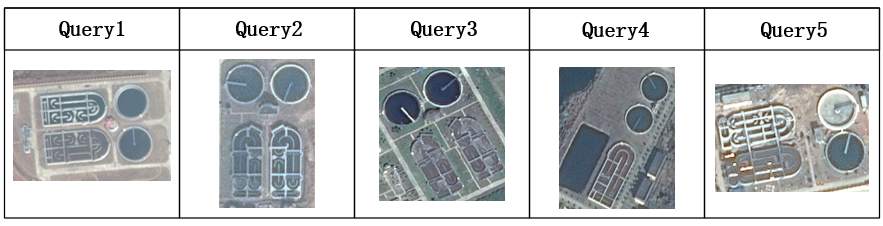}
	\end{center}
	\caption{The 5 query images of the sewage treatment plant.}
	\label{fig3}
\end{figure}\label{figure3}

\section{Method}
\label{sec:pagestyle}

The goal of this work is to detect geospatial objects on a large target image with only a single query image of a geospatial object. We need to take into account three specific issues. How to extract the image features? How to find the similar regions? How to get the objects’ bounding boxes? We illustrate these three questions in turn. The entire framework is shown in Figure \ref{fig1}. 
	\subsection{Feature extraction}
	Considering the good representation ability of deep convolutional neural networks (DCNNs), we employ VGG16 \cite{vgg} as the feature extractor and train it on the NWPU-RESISC45 \cite{nwpu45} dataset to equip it with the remote sensing domain knowledge. This dataset contains 31,500 images, covering 45 scene classes with 700 images in each class for remote sensing image scene classification. We use the trained model to extract the query and target features.
	\subsection{Multi-level feature fusion method}
	R-MAC (Regional Maximum Activation of Convolutions) is a feature encoding method proposed by \cite{r_mac}, which encodes a set of regions into short vectors over the feature maps of size ${\rm{W}} \times H \times C$, as shown in Figure \ref{fig2}. More concretely, R-MAC samples a lot of square regions uniformly at $L$ different scales. When scale $l$ is 1, the width and height of a region are both equal to $\min (W,H)$, and the number of regions is $m$. At every other scale, the number of regions is $l \times (l + m - 1)$ and the width and height are equal to $2 \times \min (W,H)/(l + 1)$. The overlap between contiguous regions is as close as possible to 40\%. The feature vector can be calculated at each region, and post-processed with $l$2-normalization. After that, we can sum all regional feature vectors into a single feature vector and $l$2-normalize it. 
	
	A multi-level feature fusion method is proposed to generate the discriminative and representative features of the query image. Inspired by R-MAC, we design the R-AMAC (Regional Average \& Maximum Activation of Convolutions) feature aggregation method where we compute the average and maximum values of each region. Thus, there are two feature vectors, the average and maximum feature vectors, for each region. Then we sum the features of all regions and concatenate the average and maximum vectors. We concatenate the global max- and average- pooling query feature vectors of block 1-3, and use the R-AMAC query feature vectors of block 4 and 5. Hence, we can get 5 feature vectors of the query image corresponding to 5 blocks of the VGG16 network. Target features are represented as 5 feature vectors of the concatenated average- and max- pooling features of all blocks in VGG16.
	
	\subsection{Similarity metric}
	Similarity can be measured by the convolution of the query feature vector and the target features for each convolution block. Therefore, we can get 5 score maps and up sample them to the same size as the input target image. After that, the final score map is the average of 5 up-sampled score maps.
	\subsection{2-stage detection pipeline}
	We set a $threshold = (mean score + max score) / 2$ to find similar regions. These similar regions of target images whose scores are higher than the $first \  threshold = 0.7$ are cropped and fed into feature extraction model to generate finer target features. Then, the similarity score is computed by the dot product between the query and the target feature vectors. Those regions whose scores are higher than the $second\ threshold = 0.9$ can be kept as the final detection results.

\section{Experiment}
\label{sec:typestyle}
In the experiments, we apply the proposed framework in sewage treatment plant and airport detections.
\subsection{Sewage treatment plant detection}
We randomly select 5 query images (see Figure \ref{fig3}) and evaluate the proposed pipeline in different settings. Some detection results are shown in Figure \ref{fig4}.
	\subsubsection{Feature fusion method analysis}
	In this part, we compare 7 different feature fusion methods, The detailed experiment settings are shown in Table 1 and the experiment results are shown in Table 2. Here, A\&MP denotes the concatenation of the average- and max- pooling features. GA\&MP denotes the concatenation of the global average- and max- pooling features.
	
	{\bf (a) Multi-level feature fusion method:} Similarity are computed with the query and target features of all blocks. For the query features, we employ global average- and max- pooling for block 1-3 and R-AMAC for block 4-5. And for the target features, we execute global average- and max- pooling for all blocks.
	
	{\bf (b) Block 5 features:}
	In this setting, we compute similarity with the query and target features of block 5. The query feature vectors are generated by executing the R-AMAC feature aggregation operation. The target features are the concatenation of the global average- and max- pooling features.
	
	{\bf (c) Average- \& max- pooling features:}
	In order to prove the effectiveness of the R-AMAC feature aggregation method in multi-level feature fusion method, we compare global average- and max- pooling to R-AMAC of block 4-5 for the query features.
	
	{\bf (d) Average-poling features:}
	We apply global average-pooling for the query features and average-pooling for the target features at each block.
	
	{\bf (e) Max-pooling features:}
	We apply global max-pooling for the query features and max-pooling for the target features at each block.
	
	{\bf (f) Average-poling features + R-AAC (Regional Average Activation of Convolutions) features:}
	We apply global average-pooling to block 1-3 and R-AAC to block 4-5 for the query features. The target features are processed by average-pooling.
	
	{\bf (g) Max-poling features + R-MAC features:}
	We apply global max-pooling to block 1-3 and R-MAC to block 4-5 for the query features. The target features are processed by max-pooling.

\begin{table}[]
	\centering
	\caption{The 7 different experiment settings of each block in the VGG16 network.}
	\resizebox{\linewidth}{80pt}{
	\begin{tabular}{c|c|c|c|c|c|c}
		\hline
		\multicolumn{2}{c|}{Method}   & block1 & block2 & block3 & block4 & block5 \\ \hline
		\multirow{2}{*}{(a)} & query  & GA\&MP & GA\&MP & GA\&MP & R-AMAC & R-AMAC \\ \cline{2-7} 
		& target & A\&MP  & A\&MP  & A\&MP  & A\&MP  & A\&MP  \\ \hline
		\multirow{2}{*}{(b)} & query  & -      & -      & -      & -      & R-AMAC \\ \cline{2-7} 
		& target & -      & -      & -      & -      & A\&MP  \\ \hline
		\multirow{2}{*}{(c)} & query  & GA\&MP & GA\&MP & GA\&MP & GA\&MP & GA\&MP \\ \cline{2-7} 
		& target & A\&M-P & A\&MP  & A\&MP  & A\&MP  & A\&MP  \\ \hline
		\multirow{2}{*}{(d)} & query  & GAP     &GAP     & GAP     & GAP     & GAP     \\ \cline{2-7} 
		& target & AP     & AP     & AP     & AP     & AP     \\ \hline
		\multirow{2}{*}{(e)} & query  & GMP     & GMP     & GMP     & GMP     & GMP     \\ \cline{2-7} 
		& target & MP     & MP     & MP     & MP     & MP     \\ \hline
		\multirow{2}{*}{(f)} & query  & GAP     & GAP     & GAP     & R-AAC  & R-AAC  \\ \cline{2-7} 
		& target & AP     & AP     & AP     & AP     & AP     \\ \hline
		\multirow{2}{*}{(g)} & query  & GMP     & GMP     & GMP     & R-MAC  & R-MAC  \\ \cline{2-7} 
		& target & MP     & MP     & MP     & MP     & MP     \\ \hline
	\end{tabular}
 }
	
	\label{table1}
\end{table}
		
	Experiment results show the effectiveness of the proposed feature fusion method. We can get discriminative and representative features by concatenating the global average- and max- pooling query features of block 1-3. For the higher-level query features of block 4-5, R-AMAC method can bring better performance of precison than global average- and max- pooling operation. Similarly, the concatenated average- and max- pooling target features have the better ability of representation compared to other feature fusion methods.	
	
	\subsubsection{Comparison of 1-stage and 2-stage pipelines}
	We also compare the performance between the 1-stage and 2-stage pipelines. As illustrated in Table \ref{table3}. Obviously, 2-stage pipeline has a similar performance of recall, but offers a significant improvement of precision.

\begin{figure}[t]
	\begin{center}
		\includegraphics[width=0.9\linewidth]{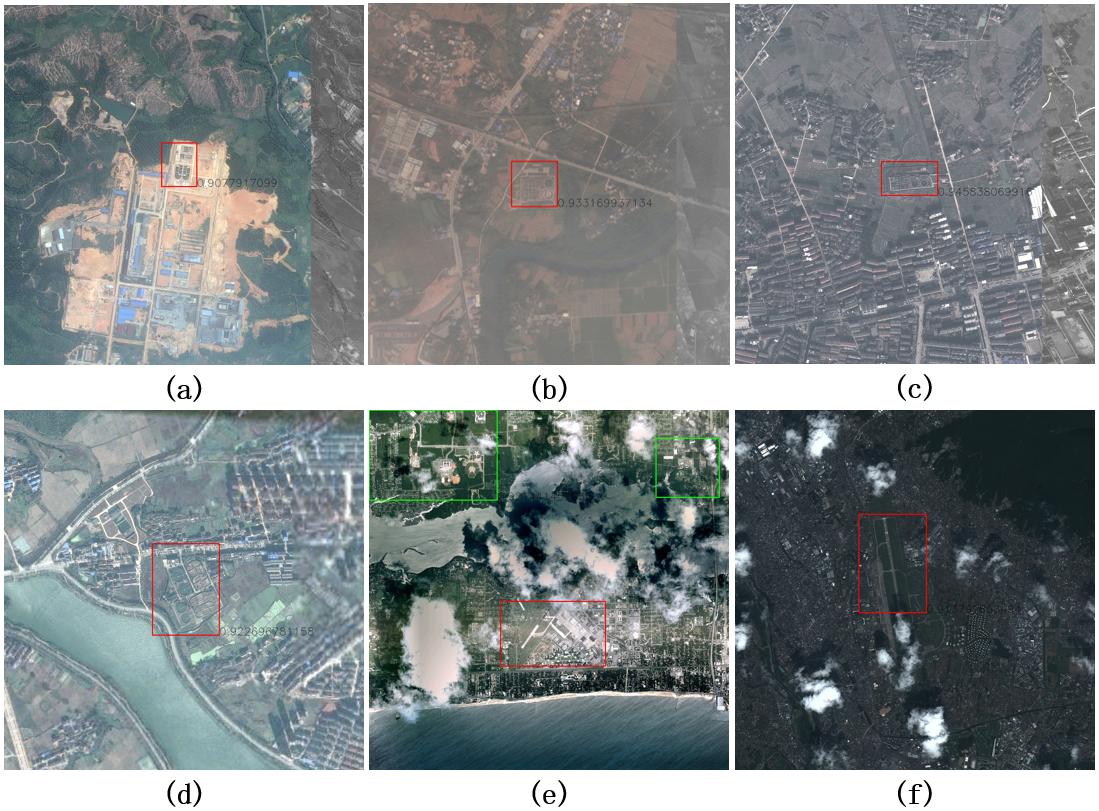}
	\end{center}
	\caption{Detection results of the proposed framework. (a)-(d) are the detection results for sewage treatment plant, (e) and (d) are the detection results for airport. In each picture, the red rectangular box denotes the region with the highest score.}
	\label{fig4}
\end{figure}\label{figure4}

\subsection{Airport detection}
We apply the proposed framework in the airport detection task. Some detection results are shown in Figure \ref{fig4}.

\begin{table}[]
	\centering
	\caption{Detection results of the 7 different experiment settings.}
		\resizebox{\linewidth}{80pt}{
			\begin{tabular}{c|c|c|c|c|c|c|c}
				\hline
				\multicolumn{2}{c|}{Method}     & query1 & query2 & query3 & query4 & query5 & mean \\ \hline
				\multirow{2}{*}{(a)} & recall   & 0.300 & 0.142 & 0.217 & 0.200 & 0.150 &0.202      \\ \cline{2-8} 
				& precision                     & 0.391 & 0.230 & 0.150 & 0.240 & 0.231 & \textbf{0.248}     \\ \hline
				\multirow{2}{*}{(b)} & recall   & 0.009 & 0.000 & 0.000 &  0.000 & 0.000 &0.002      \\ \cline{2-8} 
				& precision                     & 0.500 & 0.000 & 0.000 & 0.000 & 0.000 &0.1      \\ \hline
				\multirow{2}{*}{(c)} & recall   & 0.250 &0.217  & 0.258 & 0.192 & 0.217 &\textbf{0.227}      \\ \cline{2-8} 
				& precision                     & 0.238 &0.161  & 0.177 & 0.192 & 0.213 &0.196      \\ \hline
				\multirow{2}{*}{(d)} & recall   &0.042  &0.025  &0.067  &0.008  &0.092  &0.047      \\ \cline{2-8} 
				& precision                     &0.000  &0.000  &0.001  &0.000  &0.001  &0.000      \\ \hline
				\multirow{2}{*}{(e)} & recall   &0.133  &0.100  &0.100  &0.075  &0.142  &0.11      \\ \cline{2-8} 
				& precision                     &0.003  &0.002  &0.002  &0.002  &0.002  &0.002      \\ \hline
				\multirow{2}{*}{(f)} & recall   &0.017  &0.000  &0.033 &0.000 &0.042  &0.018      \\ \cline{2-8} 
				& precision                     &0.000  &0.000  &0.000 &0.000 &0.000  &0      \\ \hline
				\multirow{2}{*}{(g)} & recall   &0.092  &0.075  &0.058  &0.008  &0.067  &0.06      \\ \cline{2-8} 
				& \multicolumn{1}{l|}{precision} &0.001  &0.001  &0.001  &0.000  &0.001  &0.00      \\ \hline
			\end{tabular}
		}
	\label{table2}
\end{table}

\begin{table}
	\centering
	\caption{Comparison of the 1-stage and 2-stage detection pipelines.}
		\resizebox{\linewidth}{30pt}{
			\begin{tabular}{c|c|c|c|c|c|c|c}
				\hline
				\multicolumn{2}{c|}{Method}          & query1 & query2 & query3 & query4 & query5 & mean \\ \hline
				\multirow{2}{*}{1-stage} & recall    &0.358 &0.208 &0.225 &0.233 &0.233 &\textbf{0.251}      \\ \cline{2-8} 
				& precision &0.010 &0.010 &0.002 &0.010 &0.013 &0.009      \\ \hline
				\multirow{2}{*}{2-stage} & recall    &0.300 &0.142 &0.217 &0.200 &0.150 &0.202      \\ \cline{2-8} 
				& precision &0.391 &0.230 &0.150 &0.240 &0.231 &\textbf{0.248}      \\ \hline
			\end{tabular}
		}
	\label{table3}
\end{table}

\section{Conclusion}
\label{sec:majhead}

We propose a training-free, one-shot geospatial object detection framework in remote sensing images which has achieved a certain effect. The framework consistis of a feature extractor with remote sensing domain knowledge, a multi-level feature fusion method, a novel similarity metric method and a two-stage detection pipeline. Experiments on sewage treatment plant and airport detection tasks show a promising result. Besides, our method can serve as a baseline for training-free, one-shot  geospatial object detection. We will continue to extend the application scope of the framework and improve its performance in the future.

\bibliographystyle{IEEEtran}
\bibliography{egbib}

\begin{thebibliography}{1}
\providecommand{\url}[1]{#1}
\csname url@samestyle\endcsname
\providecommand{\newblock}{\relax}
\providecommand{\bibinfo}[2]{#2}
\providecommand{\BIBentrySTDinterwordspacing}{\spaceskip=0pt\relax}
\providecommand{\BIBentryALTinterwordstretchfactor}{4}
\providecommand{\BIBentryALTinterwordspacing}{\spaceskip=\fontdimen2\font plus
\BIBentryALTinterwordstretchfactor\fontdimen3\font minus
  \fontdimen4\font\relax}
\providecommand{\BIBforeignlanguage}[2]{{%
\expandafter\ifx\csname l@#1\endcsname\relax
\typeout{** WARNING: IEEEtran.bst: No hyphenation pattern has been}%
\typeout{** loaded for the language `#1'. Using the pattern for}%
\typeout{** the default language instead.}%
\else
\language=\csname l@#1\endcsname
\fi
#2}}
\providecommand{\BIBdecl}{\relax}
\BIBdecl

\bibitem{Girshick2015Fast}
R.~Girshick, ``Fast r-cnn,'' \emph{Computer Science}, 2015.

\bibitem{Ren2015Faster}
S.~Ren, K.~He, R.~Girshick, and J.~Sun, ``Faster r-cnn: towards real-time
  object detection with region proposal networks,'' in \emph{International
  Conference on Neural Information Processing Systems}, 2015, pp. 91--99.

\bibitem{model_agnostic}
C.~Finn, P.~Abbeel, and S.~Levine, ``Model-agnostic meta-learning for fast
  adaptation of deep networks,'' 2017.

\bibitem{Optimizationfewshot}
S.~Ravi and H.~Larochelle, ``Optimization as a model for few-shot learning,''
  in \emph{In International Conference on Learning Representations (ICLR)},
  2017.

\bibitem{vgg}
A.~Conneau, H.~Schwenk, L.~Barrault, and Y.~Lecun, ``Very deep convolutional
  networks for text classification,'' pp. 1107--1116, 2016.

\bibitem{nwpu45}
C.~Gong, J.~Han, and X.~Lu, ``Remote sensing image scene classification:
  Benchmark and state of the art,'' \emph{Proceedings of the IEEE}, vol. 105,
  no.~10, pp. 1865--1883, 2017.

\bibitem{r_mac}
G.~Tolias, R.~Sicre, and H.~Jégou, ``Particular object retrieval with integral
  max-pooling of cnn activations,'' \emph{Computer Science}, 2015.

\end{thebibliography}

\end{document}